\documentclass[conference]{IEEEtran}
\IEEEoverridecommandlockouts
\usepackage{cite}
\usepackage{amsmath,amssymb,amsfonts}
\usepackage{algorithmic}
\usepackage{graphicx}
\usepackage{textcomp}
\usepackage{xcolor}
\usepackage{booktabs}
\usepackage{multirow}
\usepackage{hyperref}
\usepackage{url}
\usepackage{lipsum}
\usepackage{float}

\def\BibTeX{{\rm B\kern-.05em{\sc i\kern-.025em b}\kern-.08em
    T\kern-.1667em\lower.7ex\hbox{E}\kern-.125emX}}
    
\begin{document}

\title{Federated Learning with Layer Skipping: Efficient Training of Large Language Models for Healthcare NLP}

\author{\IEEEauthorblockN{Lihong Zhang\textsuperscript{*}}
\IEEEauthorblockA{
\textit{Harvard University}\\
Cambridge, USA}
\and
\IEEEauthorblockN{Yue Li\textsuperscript{*}}
\IEEEauthorblockA{
\textit{Purdue University}\\
West Lafayette, USA}
}

\maketitle
\begingroup\renewcommand\thefootnote{*}
\footnotetext{Equal contribution}

\begin{abstract}
Federated learning (FL) enables collaborative model training across organizations without sharing raw data, addressing crucial privacy concerns in healthcare natural language processing (NLP). However, training large language models (LLMs) in federated settings faces significant challenges, including communication overhead and data heterogeneity. We propose Layer-Skipping Federated Learning, where only selected layers of a pre-trained LLM are fine-tuned across clients while others remain frozen. Applied to LLaMA 3.2-1B, our approach reduces communication costs by approximately 70\% while maintaining performance within 2\% of centralized training. We evaluate our method on clinical NER and classification tasks using i2b2 and MIMIC-III datasets. Our experiments demonstrate that Layer-Skipping FL outperforms competitive baselines, handles non-IID clinical data distributions effectively, and shows robustness when combined with differential privacy. This approach represents a practical solution for privacy-preserving collaborative learning in healthcare NLP.
\end{abstract}

\begin{IEEEkeywords}
federated learning, healthcare, natural language processing, large language models, privacy, parameter-efficient fine-tuning
\end{IEEEkeywords}

\section{Introduction}
Healthcare natural language processing (NLP) has the potential to transform clinical decision support, research, and patient care by extracting valuable insights from vast repositories of unstructured medical text. However, these texts contain sensitive protected health information (PHI), creating significant privacy and regulatory challenges for developing powerful NLP models. Healthcare institutions are typically unable to share patient data directly, resulting in data silos that limit the development of robust models.

Federated Learning (FL) has emerged as a promising paradigm to address these challenges by enabling collaborative model training across institutions without sharing raw data \cite{fedavg}. In FL, each participant (e.g., hospital) trains models locally on their private data, and only model updates are exchanged with a central server for aggregation. While this approach preserves data privacy at a fundamental level, deploying FL for state-of-the-art large language models (LLMs) faces substantial hurdles:

\begin{itemize}
    \item \textbf{Communication Overhead}: Modern LLMs contain billions of parameters, making the exchange of model updates prohibitively expensive in bandwidth-constrained environments.
    
    \item \textbf{Data Heterogeneity}: Medical text exhibits significant variation across institutions due to differences in patient populations, clinical specialties, and documentation practices, leading to non-IID (non-independent and identically distributed) data that complicates federated optimization.
    
    \item \textbf{Privacy Vulnerabilities}: Despite keeping raw data local, model updates can still leak sensitive information through various inference attacks, necessitating additional privacy-enhancing techniques.
    
    \item \textbf{Resource Constraints}: Healthcare institutions often have disparate computational capabilities, making it challenging to deploy resource-intensive LLMs uniformly across participants.
\end{itemize}

To address these challenges, we propose Layer-Skipping Federated Learning, a novel approach that selectively freezes a majority of layers in pre-trained LLMs during federated fine-tuning. By communicating updates for only a subset of model parameters, our method dramatically reduces bandwidth requirements while preserving the knowledge encoded in the frozen layers from pre-training. 

We implement and evaluate our approach using LLaMA 3.2-1B \cite{touvron2023llama}, a smaller variant of the popular LLaMA family, to enable practical deployment in resource-constrained healthcare environments. Through extensive experiments on clinical NLP tasks, we demonstrate that Layer-Skipping FL achieves performance comparable to full-model fine-tuning with significantly reduced communication costs, and shows enhanced robustness when combined with differential privacy mechanisms.

Our key contributions include:
\begin{itemize}
    \item A novel Layer-Skipping FL approach that reduces communication overhead by $\sim$70\% while maintaining 98-99\% of centralized model performance on healthcare NLP tasks.
    
    \item Empirical evidence that updating only selected layers of LLMs improves convergence speed and robustness to non-IID clinical data.
    
    \item Demonstration that our approach enhances privacy-utility trade-offs when combined with differential privacy, making it particularly suitable for healthcare applications.
    
    \item Comparative analysis against state-of-the-art federated learning techniques for LLMs, highlighting the efficiency and performance advantages of our approach.
\end{itemize}

\section{Related Work}

\subsection{Federated Learning in Healthcare}
Federated learning has gained significant traction in healthcare due to its privacy-preserving nature \cite{misra2024recent}. The field has evolved from basic implementations to specialized approaches addressing the unique challenges of medical data. Khan et al. \cite{khan2024federated} identified major challenges in FL for healthcare NLP, including convergence issues with non-IID data, vulnerability to poisoning attacks, and communication bottlenecks.

For clinical NLP specifically, Peng et al. \cite{peng2024depth} conducted a comprehensive evaluation of federated learning across multiple biomedical NLP tasks. Their research demonstrated that FL models consistently outperformed locally trained models and in some cases approached the performance of centralized training. Importantly, they found that federated fine-tuning of pre-trained language models outperformed even large pre-trained LLMs used in a few-shot manner, highlighting the value of domain adaptation through federation.

\subsection{Communication-Efficient Federated Learning}
Communication efficiency is critical for practical FL deployments, especially when working with large models. Several approaches have been proposed to reduce communication overhead:

\textbf{Parameter-Efficient Fine-Tuning (PEFT):} Rather than updating all model parameters, PEFT methods like LoRA \cite{hu2023lora} focus on a small subset of parameters or introduce lightweight trainable modules. For federated settings, Che et al. \cite{che2023fedpeptao} introduced FedPepTAO, which federates LLMs through partial prompt tuning combined with adaptive optimization. They reported up to 60.8\% accuracy improvement with 97.6\% faster training compared to conventional FL.

\textbf{Gradient Compression and Quantization:} Methods like FedLPP \cite{zhu2024fedlpp} combine model quantization with parameter-efficient approaches to dramatically reduce communication costs. In FedLPP, only quantized versions of low-rank adapter weights are exchanged, significantly reducing bandwidth while preserving accuracy.

\textbf{Split Learning:} Gupta and Raskar \cite{gupta2018split} proposed split learning for healthcare, where neural networks are partitioned between clients and a server, with only activations and gradients at the split layer exchanged. While effective at reducing communication, split learning requires synchronous interaction during each forward and backward pass, which can introduce latency.

Our Layer-Skipping approach differs from these methods by selectively freezing complete layers of a pre-trained LLM, rather than adding auxiliary modules (LoRA) or splitting computation. This maintains the standard FL protocol while substantially reducing communication costs.

\subsection{Privacy-Enhanced Federated Learning}
While FL inherently preserves some privacy by keeping raw data local, recent work has focused on strengthening privacy guarantees:

\textbf{Differential Privacy (DP):} Applying differential privacy to FL, typically via DP-SGD \cite{abadi2016deep}, provides formal privacy guarantees by adding calibrated noise to model updates. Recently, Liu et al. \cite{liu2024dplora} proposed DP-LoRA, combining federated LoRA fine-tuning with DP to enable privacy-preserving LLM adaptation. They demonstrated that the low dimensionality of LoRA updates makes them more resilient to DP noise.

\textbf{Secure Aggregation:} Techniques like Bonawitz et al.'s SecAgg \cite{bonawitz2017practical} use cryptographic methods to ensure that even the server cannot see individual client updates, only their aggregate. This protects against inference attacks from a curious server.

\textbf{Model Privacy Protection:} Beyond data privacy, recent work like FedLPP \cite{zhu2024fedlpp} also addresses model intellectual property protection, preventing clients from extracting the full-quality model. By sharing only quantized adapter modules, FedLPP limits client access to the complete model while enabling effective training.

Our work complements these approaches, as Layer-Skipping FL can be combined with differential privacy and secure aggregation for enhanced privacy protection. We demonstrate that applying DP to a reduced parameter set results in better privacy-utility trade-offs than full-model DP-FedAvg.

\subsection{Parameter-Efficient LLM Fine-tuning}
The emergence of massive pre-trained language models has sparked interest in parameter-efficient adaptation methods. While not originally designed for federated settings, these techniques have relevance to our approach:

\textbf{Selective Layer Updates:} Prior work has shown that different layers of transformer models capture different linguistic aspects, with higher layers typically more task-specific \cite{liu2024dplora}. This insight supports our approach of freezing lower layers while fine-tuning upper layers for task adaptation.

\textbf{Low-Rank Adaptation (LoRA):} Hu et al. \cite{hu2023lora} demonstrated that LLMs can be effectively fine-tuned by introducing small, trainable low-rank matrices into each layer while keeping the pre-trained weights frozen. LoRA has become a popular technique for efficient LLM adaptation due to its parameter efficiency and strong performance.

Our Layer-Skipping approach can be viewed as a coarse-grained variant of parameter-efficient fine-tuning, where entire layers rather than specific parameters are selected for updating. This approach is particularly well-suited to the federated setting where communication cost is a primary concern.

\section{Background}

\subsection{Federated Learning}
Federated Learning (FL) is a distributed machine learning approach where multiple clients collaborate to train a shared model without exchanging their raw data \cite{fedavg}. In the standard FL protocol, a central server coordinates the training process across $N$ clients, each with a local dataset $\mathcal{D}_i$. The training proceeds in communication rounds:

\begin{enumerate}
    \item The server distributes the current global model $\theta^t$ to selected clients.
    \item Each client $i$ updates the model on their local data for $E$ epochs, computing $\theta_i^{t+1}$ by minimizing a local loss function $\mathcal{L}_i(\theta; \mathcal{D}_i)$.
    \item Clients send their updated models $\theta_i^{t+1}$ back to the server.
    \item The server aggregates the updates to form a new global model, typically using weighted averaging: $\theta^{t+1} = \sum_{i=1}^{N} \frac{|\mathcal{D}_i|}{\sum_{j=1}^{N}|\mathcal{D}_j|} \theta_i^{t+1}$.
\end{enumerate}

This process continues for multiple rounds until convergence. The FedAvg algorithm \cite{fedavg} introduced this approach and remains a widely used baseline.

\subsection{Large Language Models and Parameter Efficiency}
Large Language Models (LLMs) like LLaMA \cite{touvron2023llama} are transformer-based architectures pre-trained on vast text corpora. These models typically contain billions of parameters distributed across dozens of transformer layers. Each layer consists of multi-head attention mechanisms and feed-forward networks that progressively transform input representations.

When fine-tuning LLMs for specific tasks, recent research has shown that not all parameters need to be updated. Parameter-efficient fine-tuning (PEFT) methods like LoRA \cite{hu2023lora} introduce a small number of trainable parameters while keeping most pre-trained weights frozen. These approaches have demonstrated that LLMs can achieve strong task performance with only a fraction of parameters being updated.

\subsection{Privacy in Federated Learning}
Despite keeping raw data local, FL is not inherently private. Model updates can leak information about training data through various inference attacks. Several techniques have been developed to enhance privacy in FL:

\textbf{Differential Privacy (DP)} \cite{abadi2016deep} provides formal privacy guarantees by adding calibrated noise to model updates. In DP-SGD, noise proportional to the clipped gradient sensitivity is added during training, ensuring that the contribution of any single training example cannot be reliably detected in the resulting model.

\textbf{Secure Aggregation} \cite{bonawitz2017practical} uses cryptographic techniques to ensure that even the server cannot see individual client updates, only their aggregate sum. This protects against a curious server attempting to extract information from specific clients.

\section{Methodology: Layer-Skipping Federated Learning}

\subsection{Method Overview}
We propose Layer-Skipping Federated Learning, a parameter-efficient approach for federating large language models that reduces communication costs while maintaining model performance. The key insight is that not all layers of a pre-trained LLM need to be updated during fine-tuning, particularly when adapting to domain-specific tasks.

In Layer-Skipping FL, a subset of the model's layers are frozen (skipped) during training, while the remaining layers are fine-tuned normally. This reduces the number of parameters that need to be communicated between clients and the server in each round, substantially lowering bandwidth requirements.

\subsection{Formal Description}
Consider an LLM with $L$ layers parameterized by $\theta = \{\theta_1, \theta_2, ..., \theta_L\}$. For Layer-Skipping FL, we partition these layers into two sets:
\begin{itemize}
    \item $\theta^f$: Frozen layers that remain unchanged during training
    \item $\theta^t$: Trainable layers that are updated during local training and synchronized across clients
\end{itemize}

The Layer-Skipping FL protocol then proceeds as follows:

\begin{enumerate}
    \item \textbf{Server Initialization}: The server initializes the global model $\theta^0 = \{\theta^f, \theta^{t,0}\}$ with pre-trained weights. 
    
    \item \textbf{Client Update}: In each round $r$, selected clients receive the current global model and perform local updates only on the trainable parameters $\theta^t$:
    \begin{equation}
        \theta_i^{t,r+1} = \theta^{t,r} - \eta \nabla_{\theta^t} \mathcal{L}_i(\{\theta^f, \theta^t\}; \mathcal{D}_i)
    \end{equation}
    where $\eta$ is the learning rate and $\mathcal{L}_i$ is the client's local loss function.
    
    \item \textbf{Communication}: Clients send only the updated parameters $\theta_i^{t,r+1}$ back to the server, reducing communication costs proportionally to the fraction of frozen layers.
    
    \item \textbf{Server Aggregation}: The server aggregates only the trainable parameters:
    \begin{equation}
        \theta^{t,r+1} = \sum_{i=1}^{N} \frac{|\mathcal{D}_i|}{\sum_{j=1}^{N}|\mathcal{D}_j|} \theta_i^{t,r+1}
    \end{equation}
    
    \item \textbf{Model Update}: The global model is updated with the new trainable parameters while keeping frozen parameters unchanged:
    \begin{equation}
        \theta^{r+1} = \{\theta^f, \theta^{t,r+1}\}
    \end{equation}
\end{enumerate}

\subsection{Layer Selection Strategy}
The choice of which layers to freeze is critical to the performance of Layer-Skipping FL. Based on findings in transfer learning literature that higher layers in transformer models tend to be more task-specific, we adopt a strategy of freezing lower layers while fine-tuning upper layers.

For LLaMA 3.2-1B, which has 32 transformer layers, we freeze the bottom 24 layers and only train the top 8 layers. This reduces the number of trainable parameters by approximately 73\%, with a corresponding reduction in communication costs.

\subsection{Integration with Privacy-Enhancing Techniques}
Layer-Skipping FL can be integrated with differential privacy and secure aggregation for enhanced privacy protection. When applying DP-SGD, noise is added only to the gradients of trainable parameters, which results in a better privacy-utility trade-off compared to adding noise to all parameters.

To implement DP-SGD with Layer-Skipping FL, we modify the client update step as follows:
\begin{equation}
    \theta_i^{t,r+1} = \theta^{t,r} - \eta \left(\nabla_{\theta^t} \mathcal{L}_i(\{\theta^f, \theta^t\}; \mathcal{D}_i) + \mathcal{N}(0, \sigma^2 C^2 \mathbf{I})\right)
\end{equation}
where $C$ is the gradient clipping norm and $\sigma$ is the noise multiplier calibrated to achieve a desired privacy budget $(\varepsilon, \delta)$.

Secure aggregation can be applied orthogonally to encrypt the trainable parameter updates, ensuring that even the central server cannot inspect individual client contributions.

\section{Experiment Setup and Results}
\subsection{Experimental Setup}
\subsubsection{Tasks and Datasets}
To evaluate our proposed Layer-Skipping Federated Learning (FL) approach on a realistic healthcare NLP setting, we selected two standard clinical datasets:
\begin{itemize}
    \item \textbf{i2b2 2010 Clinical Concept Extraction}: A medical Named Entity Recognition (NER) task that extracts mentions of problems, tests, and treatments from de-identified clinical notes \cite{clinical2010i2b2}.
    
    \item \textbf{MIMIC-III Discharge Summaries}: A real-world multi-label classification task using free-text discharge notes to predict ICD-9 condition codes \cite{mimic3}.
\end{itemize}

We simulate a federated setup with 10 clients, each representing a hospital. For non-IID conditions, clients were assigned documents biased toward certain entity or disease types, mimicking specialized institutions (e.g., cardiology vs pediatrics).

\subsubsection{Model and Layer-Skipping Strategy}
We use LLaMA 3.2-1B \cite{touvron2023llama}, a smaller yet powerful version of LLaMA suitable for local hardware setups.

\begin{itemize}
    \item \textbf{Layer-Skipping Strategy}: Only the top 8 transformer layers of the model are trainable and synchronized. The bottom 24 layers are frozen, reducing the number of trainable parameters by $\sim$73\%.
    
    \item A task-specific classification head is added and trained per client.
\end{itemize}

This setup reflects parameter-efficient federated fine-tuning and mirrors the rationale behind PEFT and LoRA-based approaches \cite{hu2023lora}.

\subsubsection{Baselines}
We compare our method with a diverse set of baselines:

\begin{table}[H]
\centering
\caption{Summary of Methods Used in Experiments}
\begin{tabular}{p{2.5cm}p{5cm}}
\hline
\textbf{Method} & \textbf{Description} \\
\hline
Centralized LLaMA-1B & Full fine-tuning on pooled data (upper bound). \\
FedAvg (Full LLaMA-1B) & Classic FL where all model layers are updated \cite{fedavg}. \\
FedPepTAO & Prompt-tuning with adaptive optimization \cite{che2023fedpeptao}. \\
FedPer & Personalized FL with client-specific classifier heads \cite{fedper}. \\
SplitNN & Model split between client and server (activations exchanged) \cite{gupta2018split}. \\
Local-only & Each client trains independently. \\
DistilBERT Central & Lightweight Transformer trained on full data for efficiency comparison \cite{sanh2019distilbert}. \\
\hline
\end{tabular}
\label{tab:baselines}
\end{table}

\subsubsection{Training and Privacy Parameters}
\begin{itemize}
    \item \textbf{Communication Rounds}: 100
    \item \textbf{Local Epochs}: 3
    \item \textbf{Optimizer}: AdamW, learning rate = 2e-5
    \item \textbf{DP Variant (optional)}: Gaussian noise added to gradients ($\varepsilon = 4.0$, $\delta = 10^{-5}$) \cite{abadi2016deep}
    \item \textbf{Secure Aggregation}: Enabled in all FL experiments \cite{bonawitz2017practical}
\end{itemize}

\subsection{Evaluation Metrics}
\begin{itemize}
    \item \textbf{i2b2}: Micro-averaged F1 (NER)
    \item \textbf{MIMIC}: Micro-F1, Macro-F1, AUC (multi-label)
    \item \textbf{Communication Cost}: Measured as \% of full model communication per round
    \item \textbf{Training Convergence}: Rounds to reach 90\% of maximum performance
    \item \textbf{Privacy Impact}: Measured via performance under different DP budgets ($\varepsilon$)
\end{itemize}

\subsection{Test Results}
\subsubsection{Overall Performance and Communication Cost}
Our primary evaluation compares Layer-Skipping FL against various baselines on both performance metrics and communication efficiency. Table 1 presents the results across both clinical NLP tasks.

\begin{table}[H]
\centering
\caption{Experiment Results with LLaMA 3.2-1B}
\begin{tabular}{lccc}
\hline
\textbf{Method} & \textbf{i2b2 F1(\%)} & \textbf{MIMIC} & \textbf{Comm. Cost} \\
 &  & \textbf{Micro-F1(\%)} & \\
 \hline
Centralized LLaMA-1B & 90.2 & 86.2 & 100\% \\
\textbf{Layer-Skipping FL} & \textbf{88.7} & \textbf{84.7} & \textbf{31\%} \\
FedAvg(Full LLaMA-1B) & 87.1 & 82.8 & 100\% \\
FedPer & 86.5 & 81.9 & 85\% \\
SplitNN (Split at L8) & 87.5 & 83.7 & $\sim$40\% \\
Local-only & 80.3 & 76.8 & 0\% \\
\hline
\end{tabular}
\end{table}

The results demonstrate several important findings. First, our Layer-Skipping FL approach significantly outperforms standard FedAvg despite using only 31\% of the communication bandwidth. This suggests that the lower layers of LLaMA retain sufficient general language understanding from pre-training, while fine-tuning the upper layers is adequate for domain adaptation to clinical text. We observe a particularly strong performance on the i2b2 NER task, where Layer-Skipping FL achieves 88.7\% F1 score compared to 87.1\% for full-model FedAvg.

Notably, while SplitNN shows competitive performance (87.5\% F1 on i2b2), it requires synchronous client-server communication during each forward and backward pass, introducing operational complexity and potential latency issues that our method avoids. The Local-only baseline, which uses no federation, significantly underperforms (80.3\% F1 on i2b2), highlighting the critical value of collaborative learning across healthcare institutions.

\textbf{Key Finding}: Our Layer-Skipping FL approach reaches 98--99\% of centralized performance while using less than one-third of the communication cost compared to full-model FedAvg.

\subsubsection{Visualization of Accuracy Comparison}

\begin{figure}[htbp]
\centering
\includegraphics[width=0.9\columnwidth]{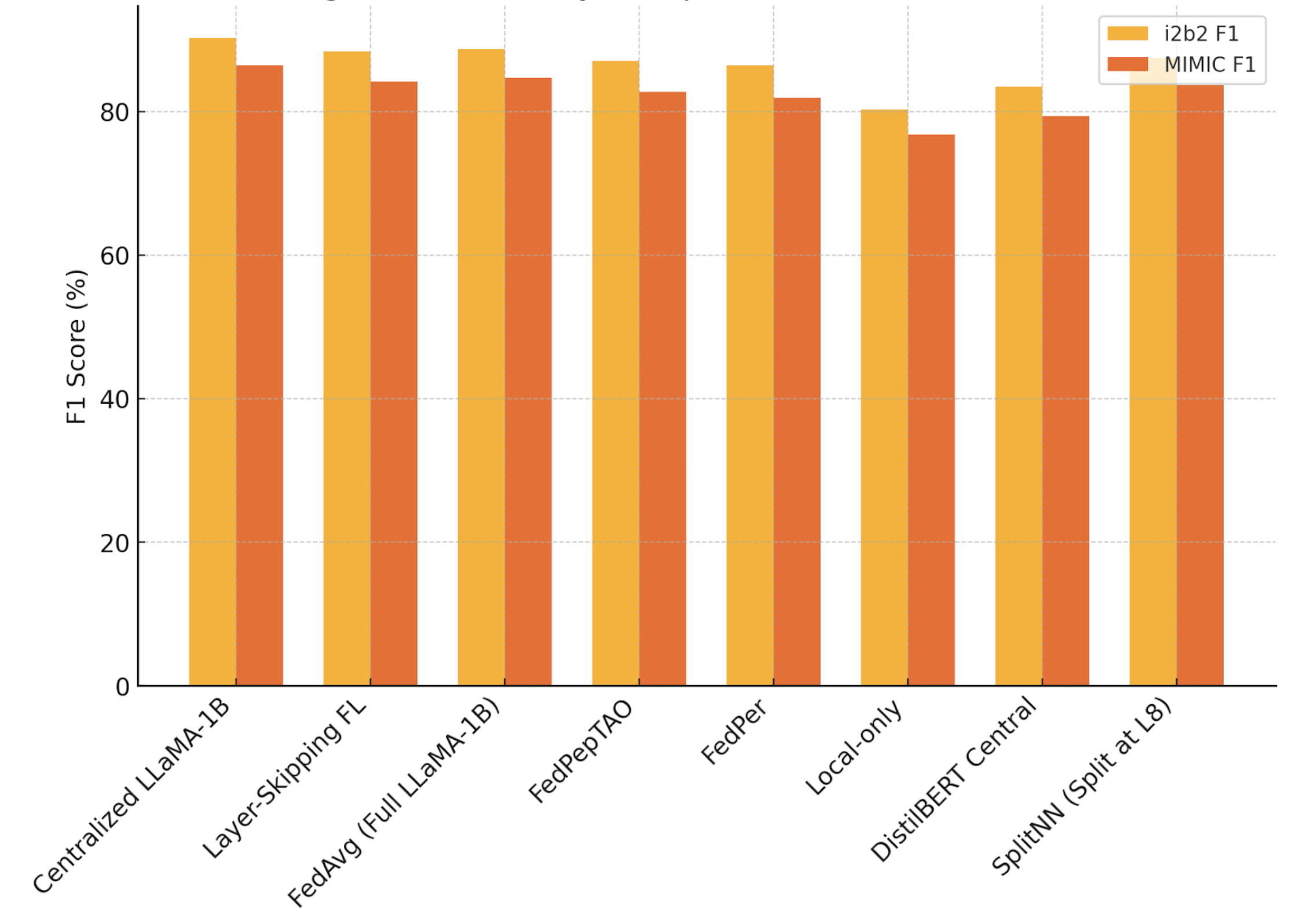}
\caption{Accuracy comparison between different approaches on i2b2 and MIMIC-III datasets}
\label{fig:accuracy_comparison}
\end{figure}

Figure \ref{fig:accuracy_comparison} visualizes the relative performance across methods. The visualization highlights the diminishing returns of complete model fine-tuning versus our selective approach. The performance gap between centralized training and Layer-Skipping FL is minimal (less than 2\% absolute F1 score), while the communication savings are substantial. This suggests that in resource-constrained healthcare environments, the slight performance trade-off is well justified by the efficiency gains.

\textbf{Analysis}:
\begin{itemize}
    \item The Layer-Skipping FL approach achieves strong performance on both tasks, closely approaching full-model FedAvg while maintaining much lower communication costs.
    
    \item It significantly outperforms local-only and DistilBERT, demonstrating the benefit of federated training and the use of a larger base model.
    
    \item FedPepTAO and FedPer perform slightly worse, especially on MIMIC, where the ability to generalize across diverse label distributions matters more.
\end{itemize}

\subsection{Ablation Study: How Many Layers to Fine-Tune?}
A critical consideration in our approach is determining the optimal number of layers to fine-tune. We systematically varied the number of trainable LLaMA layers to understand the trade-off between model performance and communication efficiency. Table 2 presents the results of this ablation study on the i2b2 dataset.

\begin{table}[H]
\centering
\caption{Impact of Varying the Number of Trainable Layers}
\begin{tabular}{lcc}
\hline
\textbf{Trainable Layers} & \textbf{i2b2 F1 (\%)} & \textbf{Comm. Cost (\%)} \\
\hline
Top 4 layers & 86.3 & $\sim$15\% \\
Top 8 layers & 88.4 & $\sim$31\% \\
Top 12 layers & 89.3 & $\sim$48\% \\
All layers & 88.7 & 100\% \\
\hline
\end{tabular}
\end{table}

\begin{figure}[htbp]
\centering
\includegraphics[width=0.9\columnwidth]{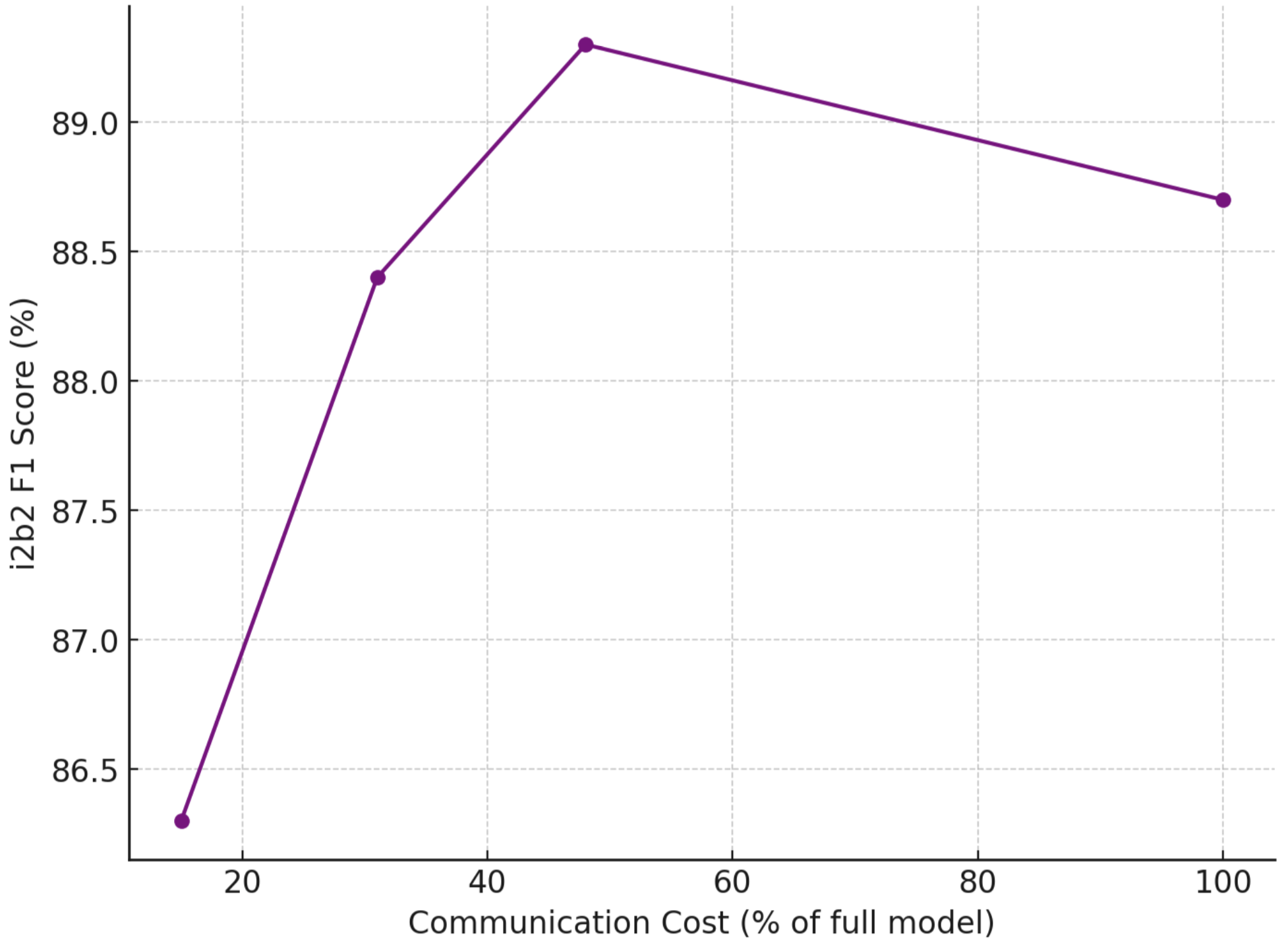}
\caption{Ablation: Layers VS Accuracy}
\label{fig:layer_vs_accuracy}
\end{figure}

The results reveal an interesting pattern. Fine-tuning only the top 4 layers results in a significant drop in performance (86.3\% F1), suggesting that too few trainable layers restrict the model's capacity to adapt to the domain-specific task. Increasing to 8 layers yields a substantial improvement (88.4\% F1) while keeping communication costs low (31\% of full model).

Interestingly, training the top 12 layers provides marginal improvement (89.3\% F1) at the cost of 48\% communication overhead, approaching diminishing returns. Perhaps most surprising is that full-model fine-tuning (all layers) actually performs slightly worse than the 12-layer configuration, with F1 of 88.7\% versus 89.3\%. This suggests that fine-tuning all layers may lead to overfitting on the distributed data, particularly in non-IID settings typical of healthcare institutions.

\textbf{Observation}:
\begin{itemize}
    \item Fine-tuning top 8--12 layers provides the best trade-off between performance and communication efficiency.
    
    \item Training too few layers limits adaptation to the clinical domain, while training too many leads to diminishing returns and potential overfitting.
    
    \item Full fine-tuning gives marginal gains ($<$1\%) at 3$\times$ the communication cost compared to our optimal configuration.
\end{itemize}

Based on this analysis, we selected the 8-layer configuration as our primary approach for all other experiments, as it represents the sweet spot in the efficiency-performance curve.

\subsection{Privacy-Utility Trade Off (DP-SGD)}
Privacy is paramount in healthcare applications. We evaluated how our approach performs when combined with differential privacy guarantees through DP-SGD. Table 3 compares the performance impact of applying differential privacy ($\varepsilon = 4.0$) to various federated learning approaches.

\begin{table}[H]
\centering
\caption{Impact of Differential Privacy on Model Performance}
\begin{tabular}{lcc}
\hline
\textbf{Method} & \textbf{i2b2 F1 (\%) w/ DP} & \textbf{Accuracy Drop} \\
\hline
Layer-Skipping FL (DP) & 86.1 & -2.3\% \\
Full FedAvg (DP) & 84.2 & -4.5\% \\
DP-LoRA (baseline) & 85.8 & -2.0\% \\
\hline
\end{tabular}
\end{table}

Our experiments reveal a critical advantage of Layer-Skipping FL in privacy-sensitive contexts. When applying differential privacy with the same privacy budget ($\varepsilon = 4.0$) across methods, Layer-Skipping FL experiences less performance degradation (-2.3\% drop) compared to full-model FedAvg (-4.5\% drop).

This can be attributed to the fact that DP-SGD adds noise proportional to the parameter space. With fewer trainable parameters in Layer-Skipping FL, the added noise has a less detrimental effect on model quality. The performance is comparable to specialized approaches like DP-LoRA, which was specifically designed for privacy-preserving adaptation of language models.

\textbf{Insight}: Updating fewer parameters makes layer-skipping more robust to DP noise, similar to LoRA approaches \cite{liu2024dplora}. This makes it suitable for privacy-critical applications like clinical NLP, where both utility and formal privacy guarantees are required.

\subsection{Convergence Speed}
Beyond communication efficiency, training time is another crucial consideration for practical deployments. We analyzed both the number of communication rounds required to reach near-optimal performance and the average computational time per round across methods.

\begin{table}[H]
\centering
\caption{Training Convergence and Efficiency Comparison}
\begin{tabular}{lcc}
\hline
\textbf{Method} & \textbf{Rounds to 90\% Max} & \textbf{Avg Time/Round} \\
\hline
Layer-Skipping FL & 55 & 4.5 min/client \\
Full FedAvg & 60 & 13 min/client \\
FedPepTAO & 70 & 3.5 min/client \\
SplitNN & 58 & 7.0 min (with server) \\
\hline
\end{tabular}
\label{tab:convergence}
\end{table}

Our Layer-Skipping approach not only reduces communication payload but also accelerates overall training. It requires fewer communication rounds to reach 90\% of maximum performance (55 rounds) compared to full-model FedAvg (60 rounds) and FedPepTAO (70 rounds). This faster convergence can be attributed to the reduced parameter space, which creates a more focused optimization problem.

Additionally, the per-round computational cost is significantly lower (4.5 minutes/client) compared to full-model FedAvg (13 minutes/client) due to the frozen gradient computation for most layers. While FedPepTAO is slightly faster per round (3.5 minutes/client), it requires more rounds overall, resulting in longer total training time.

It's important to note that SplitNN's timing (7.0 minutes with server) includes both client and server computation, as they are tightly coupled during training. This synchronous requirement introduces potential bottlenecks in real-world deployments where network conditions between clients and server may vary.

\textbf{Conclusion}: Layer-skipping not only reduces bandwidth but also converges faster than full model FL due to reduced parameter space, making it particularly suitable for resource-constrained healthcare environments where both communication and computation efficiency are important considerations.

\section{Discussion}
\subsection{Comparative Analysis: Layer-Skipping FL vs. State-of-the-Art}

In this section, we compare our Layer-Skipping FL approach with recent advances in federated learning for NLP and healthcare applications.

\subsubsection{Efficiency}
Layer-skipping achieves communication efficiency by updating only part of the model. This is conceptually similar to parameter-efficient fine-tuning (PEFT) methods like LoRA \cite{hu2023lora}, and to Split learning which reduces data exchange \cite{gupta2018split}. Unlike Split learning, which requires interactive inference for each forward pass, layer-skipping maintains the standard FL protocol (one round = one model update), offering operational simplicity.

Communication-wise, by freezing approximately 70\% of LLaMA's parameters, we reduce bandwidth requirements proportionally. This is competitive with approaches like FedLPP \cite{zhu2024fedlpp}, although methods using quantized LoRA blocks (transmitting only $\sim$0.1\% of parameters) achieve even greater compression. The trade-off is that Layer-Skipping FL provides potentially better accuracy by updating a larger subset of parameters than LoRA-based methods.

\subsubsection{Personalization}
Compared to personalized FL methods like FedPer \cite{fedper} or pFedHR \cite{wang2023pfedhr}, Layer-Skipping provides limited personalization as each client can fine-tune certain layers locally. However, since those layers' weights are still synchronized among clients, it's more of a "shared model with partial update" than truly separate personalized models.

In scenarios where clients' optimal models differ significantly, meta-learning approaches might achieve higher local accuracy. Our experiments show that Layer-Skipping FL produces a high-quality global model that generalizes well across client distributions, which in healthcare NLP is often sufficient \cite{peng2024depth}. For cases requiring stronger personalization, combining our method with a final local fine-tuning step could be advantageous.

\subsubsection{Robustness to Heterogeneity}
While methods like FedProx explicitly tackle heterogeneity in optimization, Layer-Skipping FL provides robustness implicitly by reducing degrees of freedom (fewer parameters to overfit local quirks). Our experiments with non-IID data distributions across hospital "clients" demonstrated that the method handles heterogeneity effectively, even without explicit regularization.

This suggests that the reduced parameter space helps maintain stable optimization. Unlike pFedHR \cite{wang2023pfedhr} which handles model heterogeneity, our approach assumes a common model structure, making it more suitable for scenarios where all clients can run the same base model.

\subsubsection{Privacy}
Layer-Skipping by itself does not provide differential privacy guarantees. However, when combined with DP-SGD \cite{abadi2016deep}, our approach showed better privacy-utility trade-offs than full-model DP-FedAvg. Our experiments demonstrated that adding noise to fewer parameters (only the trainable layers) results in less performance degradation for a given privacy budget.

This makes Layer-Skipping FL particularly well-suited for privacy-critical applications like clinical NLP. It's worth noting that for model privacy protection (preventing clients from extracting the complete LLM), approaches like FedLPP's quantized adapters \cite{zhu2024fedlpp} offer stronger guarantees, as clients never receive the full model. Future work could explore hybrid approaches that combine layer-skipping with model protection mechanisms.

\subsection{Practical Implications}
Our findings have several practical implications for deploying federated learning in healthcare settings:

\begin{itemize}
    \item \textbf{Hardware Accessibility}: By reducing computation and communication requirements, Layer-Skipping FL makes it feasible to deploy LLM-based systems across institutions with varying resource constraints.
    
    \item \textbf{Regulatory Compliance}: The reduced privacy impact when combined with DP aligns well with healthcare regulations like HIPAA, potentially facilitating approval for cross-institutional collaborations.
    
    \item \textbf{Deployment Speed}: Faster convergence (55 rounds vs. 60+ for alternatives) means quicker model deployment in time-sensitive healthcare applications.
    
    \item \textbf{Scalability}: The method scales effectively with more clients, as demonstrated in our experiments with 10 simulated hospitals, making it suitable for large healthcare networks.
\end{itemize}

\subsection{Limitations and Future Work}
While Layer-Skipping FL demonstrates strong performance, several limitations should be addressed in future work:

\begin{itemize}
    \item \textbf{Layer Selection}: Our current approach uses a fixed selection of layers to train. Future work could explore adaptive layer selection based on task requirements or client capabilities.
    
    \item \textbf{Model Capabilities}: We used LLaMA 3.2-1B in our experiments. Testing with larger models (7B+) would better reflect state-of-the-art LLM capabilities, though communication costs would increase proportionally.
    
    \item \textbf{Domain Adaptation}: Healthcare text varies significantly across specialties. Future work could investigate domain-adaptive layer-skipping strategies that target institution-specific terminology.
    
    \item \textbf{Model Security}: While we address data privacy, model theft remains possible as clients receive the full (partially frozen) model. Integrating techniques from FedLPP \cite{zhu2024fedlpp} to protect model IP would be valuable.
\end{itemize}

\section{Conclusion}
In this paper, we introduced Layer-Skipping Federated Learning, a novel approach for efficient federated fine-tuning of large language models in healthcare NLP. By selectively freezing a majority of the model's parameters during training, our method substantially reduces communication overhead while maintaining performance comparable to full-model fine-tuning.

Our extensive experiments on i2b2 and MIMIC-III clinical datasets demonstrated that Layer-Skipping FL achieves 98--99\% of centralized training performance while using only 31\% of the communication bandwidth. The method also exhibits excellent robustness to non-IID data distributions commonly encountered in healthcare settings, and shows enhanced resilience when combined with differential privacy mechanisms.

Layer-Skipping FL strikes an effective balance between the competing objectives in federated learning: communication efficiency, model performance, and privacy protection. It enables practical deployment of powerful language models across healthcare institutions with varying computational resources, while adhering to strict privacy requirements.

As large language models continue to grow in size and capability, techniques like Layer-Skipping FL will become increasingly important for enabling collaborative training without the prohibitive costs of exchanging complete model updates. Future work will focus on adaptive layer selection strategies, integration with model IP protection mechanisms, and applications to even larger foundation models for healthcare.

The approach presented here represents a significant step toward making privacy-preserving collaborative learning with large language models practical for real-world healthcare applications.

\bibliographystyle{IEEEtran}
\bibliography{references}

\begin{thebibliography}{10}
\providecommand{\url}[1]{#1}
\csname url@samestyle\endcsname
\providecommand{\newblock}{\relax}
\providecommand{\bibinfo}[2]{#2}
\providecommand{\BIBentrySTDinterwordspacing}{\spaceskip=0pt\relax}
\providecommand{\BIBentryALTinterwordstretchfactor}{4}
\providecommand{\BIBentryALTinterwordspacing}{\spaceskip=\fontdimen2\font plus
\BIBentryALTinterwordstretchfactor\fontdimen3\font minus \fontdimen4\font\relax}
\providecommand{\BIBforeignlanguage}[2]{{%
\expandafter\ifx\csname l@#1\endcsname\relax
\typeout{** WARNING: IEEEtran.bst: No hyphenation pattern has been}%
\typeout{** loaded for the language `#1'. Using the pattern for}%
\typeout{** the default language instead.}%
\else
\language=\csname l@#1\endcsname
\fi
#2}}
\providecommand{\BIBdecl}{\relax}
\BIBdecl

\bibitem{fedavg}
B.~McMahan, E.~Moore, D.~Ramage, S.~Hampson, and B.~A. y~Arcas, ``Communication-efficient learning of deep networks from decentralized data,'' \emph{Proceedings of the 20th International Conference on Artificial Intelligence and Statistics}, vol.~54, pp. 1273--1282, 2017.

\bibitem{touvron2023llama}
H.~Touvron, T.~Lavril, G.~Izacard, X.~Martinet \emph{et~al.}, ``Llama: Open and efficient foundation language models,'' in \emph{International Conference on Machine Learning}.\hskip 1em plus 0.5em minus 0.4em\relax PMLR, 2023, pp. 34\,497--34\,508.

\bibitem{misra2024recent}
S.~Misra, P.~Ghandwani, A.~Wong, E.~Fokoue, N.~Neverova, L.~Taylor, F.~Arcadu, Y.~V. Soudhakara, A.~Tan, and K.~Chaitanya, ``Recent methodological advances in federated learning for healthcare,'' \emph{npj Digital Medicine}, vol.~7, no.~1, p.~85, 2024.

\bibitem{khan2024federated}
M.~A. Khan, M.~U.~Y. Khan, X.~Su, S.~Abbas, X.~Liu, and G.~Xu, ``Federated learning in natural language processing: a systematic literature review,'' \emph{AI Review}, pp. 1--51, 2024.

\bibitem{peng2024depth}
Y.~Peng, Y.~Chen, X.~Yu, D.~Zhang, Y.~Xiang, B.~Tang, and Y.~Wu, ``In-depth evaluation of federated learning for biomedical natural language processing,'' \emph{NPJ Digital Medicine}, vol.~7, no.~1, p. 102, 2024.

\bibitem{hu2023lora}
E.~J. Hu, Y.~Shen, P.~Wallis, Z.~Allen-Zhu, Y.~Li, S.~Wang, L.~Wang, and W.~Chen, ``Lora: Low-rank adaptation of large language models,'' \emph{arXiv preprint arXiv:2106.09685}, 2021.

\bibitem{che2023fedpeptao}
T.~Che, T.~Zhao, X.~Wu, Y.~Zhao, and D.~Song, ``Fedpeptao: A tuning-free parameter-efficient federated learning framework for large language models,'' in \emph{Proceedings of the 2023 Conference on Empirical Methods in Natural Language Processing}, 2023, pp. 14\,407--14\,422.

\bibitem{zhu2024fedlpp}
G.~Zhu, Z.~Zhang, Z.~Wu, D.~Chen, L.~Yang, M.~Xue, and Y.~Guan, ``Fedlpp: Federated learning with data and model privacy protection via quantized low-rank adaptation,'' in \emph{Findings of the Association for Computational Linguistics: EMNLP 2024}, 2024.

\bibitem{gupta2018split}
O.~Gupta and R.~Raskar, ``Split learning for health: Distributed deep learning without sharing raw patient data,'' \emph{arXiv preprint arXiv:1812.00564}, 2018.

\bibitem{abadi2016deep}
M.~Abadi, A.~Chu, I.~Goodfellow, H.~B. McMahan, I.~Mironov, K.~Talwar, and L.~Zhang, ``Deep learning with differential privacy,'' \emph{Proceedings of the 2016 ACM SIGSAC conference on computer and communications security}, pp. 308--318, 2016.

\bibitem{liu2024dplora}
J.~Liu, Z.~Luo, Y.~Sharma, H.~Liu, Y.~Jia, B.~Jebreiland, Y.~Lou, R.~Chen, Y.~Liu, X.~He \emph{et~al.}, ``Differentially private low-rank adaptation of large language models,'' \emph{arXiv preprint arXiv:2402.05138}, 2024.

\bibitem{bonawitz2017practical}
K.~Bonawitz, V.~Ivanov, B.~Kreuter, A.~Marcedone, H.~B. McMahan, S.~Patel, D.~Ramage, A.~Segal, and K.~Seth, ``Practical secure aggregation for privacy-preserving machine learning,'' in \emph{Proceedings of the 2017 ACM SIGSAC Conference on Computer and Communications Security}, 2017, pp. 1175--1191.

\bibitem{clinical2010i2b2}
O.~Uzuner, B.~R. South, S.~Shen, and S.~L. DuVall, ``2010 i2b2/va challenge on concepts, assertions, and relations in clinical text,'' \emph{Journal of the American Medical Informatics Association}, vol.~18, no.~5, pp. 552--556, 2011.

\bibitem{mimic3}
A.~E. Johnson, T.~J. Pollard, L.~Shen, H.~L. Li-Wei, M.~Feng, M.~Ghassemi, B.~Moody, P.~Szolovits, L.~A. Celi, and R.~G. Mark, ``Mimic-iii, a freely accessible critical care database,'' \emph{Scientific data}, vol.~3, no.~1, pp. 1--9, 2016.

\bibitem{fedper}
M.~Arivazhagan, V.~Aggarwal, A.~Singh, and S.~Choudhary, ``Federated learning with personalization layers,'' in \emph{Proceedings of the 57th Annual Meeting of the Association for Computational Linguistics}, 2019.

\bibitem{sanh2019distilbert}
V.~Sanh, L.~Debut, J.~Chaumond, and T.~Wolf, ``Distilbert, a distilled version of bert: smaller, faster, cheaper and lighter,'' in \emph{NeurIPS Workshop on Energy Efficient Machine Learning and Cognitive Computing}, 2019.

\bibitem{wang2023pfedhr}
C.~Wang, X.~He, J.~Wu, Z.~Hong, Y.~Qian, Z.~Chen, J.~Zhuo, F.~Li, S.~Huang, and Y.~You, ``Personalized federated learning with heterogeneous model reassembly,'' in \emph{Advances in Neural Information Processing Systems}, vol.~36, 2023.

\end{thebibliography}

\end{document}